# The Dynamic Controllability of Conditional STNs with Uncertainty

**Luke Hunsberger***
hunsberg@cs.vassar.edu
Vassar College
Poughkeepsie, NY 12604

**Roberto Posenato**
roberto.posenato@univr.it
University of Verona
Verona, Italy

**Carlo Combi**
carlo.combi@univr.it
University of Verona
Verona, Italy

### Abstract

Recent attempts to automate business processes and medical-treatment processes have uncovered the need for a formal framework that can accommodate not only temporal constraints, but also observations and actions with uncontrollable durations. To meet this need, this paper defines a Conditional Simple Temporal Network with Uncertainty (CSTNU) that combines the simple temporal constraints from a Simple Temporal Network (STN) with the conditional nodes from a Conditional Simple Temporal Problem (CSTP) and the contingent links from a Simple Temporal Network with Uncertainty (STNU). A notion of dynamic controllability for a CSTNU is defined that generalizes the dynamic consistency of a CTP and the dynamic controllability of an STNU. The paper also presents some sound constraint-propagation rules for dynamic controllability that are expected to form the backbone of a dynamic-controllability-checking algorithm for CSTNUs.

## Introduction and Motivation

Workflow systems have been used to model business, manufacturing and medical-treatment processes. However, as Bettini et al. (2002) observed: "It would greatly enhance the capabilities of current workflow systems if quantitative temporal constraints on the duration of activities and their synchronization requirements can be specified and reasoned about." Toward that end, Combi et al. (2007; 2009; 2010) presented a new workflow model that accommodates the following key features: tasks with uncertain/uncontrollable durations; temporal constraints among tasks; and branching paths, where the branch taken is not known in advance. Fig. 1 shows a sample workflow from the health-care domain, similar to one presented by Combi and Posenato (2009). In this workflow, all times are in minutes, and:

- tasks are represented by rounded boxes;
- branching points are represented by nine-sided boxes called *split* or *join connectors*[1];
- tasks and connectors have duration attributes, $[x, y]$;

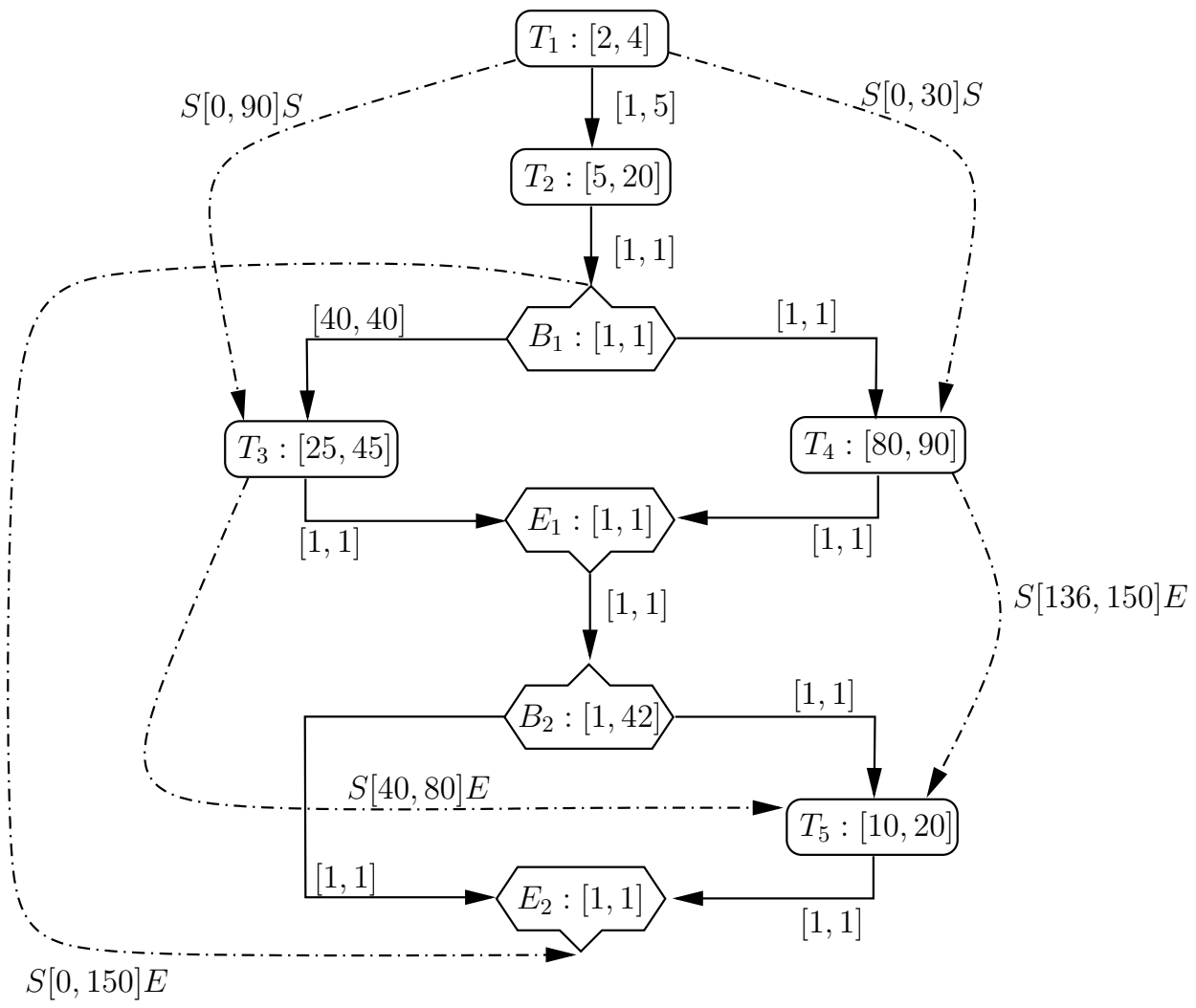

*Figure 1:* A sample workflow

- the flow, represented by solid arrows moving downward, specifies a partial order among tasks and connectors;
- intervals between consecutive tasks or connectors—called *delays*—are bounded by intervals of the form $[x, y]$;
- additional temporal constraints are represented by dashed arrows, also labeled by intervals of the form $[x, y]$.

The $S$ and $E$ notations on temporal constraints are used to indicate whether a constraint applies to the *starting* or *ending* times of the tasks/connectors it links. For example, the notation $S[136, 150]E$ on the arrow from $T_4$ to $T_5$ indicates that the duration of the interval from the *start* of $T_4$ to the *end* of $T_5$ must be in the range, $[136, 150]$.

The tasks and their uncontrollable durations are:

$T_1$: Pre-hospital issues, 2–4 min.

$T_2$: Initial patient evaluation, 5–20 min.

$T_3$: Percutaneous Coronary Intervention, 25–45 min.

$T_4$: Reperfusion Fibrinolytic therapy, 80–90 min.

$T_5$: Ancillary therapy, 10–20 min.

The semantics of execution for workflows stipulates that:

*Funded in part by the Phoebe H. Beadle Science Fund.

[1] Combi and Posenato (2009) used diamonds for connectors.

- The agent is free to choose starting times for all tasks, but does not control their durations; instead, the agent merely observes the durations of tasks in real time.
- The agent is free to choose starting and ending times for all connectors, but does not control which branch a *split* connector will follow; instead, the agent merely observes which branch is followed in real time.[2]

If a workflow admits a strategy for executing its tasks and connectors such that

- its execution decisions depend only on past observations of task durations and branch directions; and
- all delays and temporal constraints are guaranteed to be satisfied no matter how the task durations turn out and no matter which branches are taken, then that workflow is said to be *history-dependent controllable* (Combi and Posenato 2009; 2010). The workflow in Fig. 1 is history-dependent controllable. A successful strategy must restrict the interval for $B_2$ to be $[32, 42]$ if the branch containing task $T_4$ was taken, or $[1, 31]$ if the branch containing task $T_3$ was taken. Combi and Posenato (2010) presented an exponential-time algorithm for determining whether any workflow is history-dependent controllable.

The rest of this paper introduces a Conditional Simple Temporal Network with Uncertainty (CSTNU) that provides a formal representation for the time-points and temporal constraints underlying workflows. A CSTNU is shown to generalize three existing kinds of temporal networks from the literature. Similarly, the concept of dynamic controllability for a CSTNU is shown to generalize analogous concepts for existing kinds of networks, while also being related to the history-dependent controllability for a workflow.

## Related Temporal Networks

This section summarizes three kinds of temporal networks from the literature: Simple Temporal Networks, Conditional Simple Temporal Problems, and Simple Temporal Networks with Uncertainty. For convenience, we replace the Conditional Simple Temporal Problem with an equivalent alternative: the Conditional Simple Temporal Network.

### Simple Temporal Networks

**Definition 1** (STN). (Dechter, Meiri, and Pearl 1991) A *Simple Temporal Network* (STN) is a pair, $(\mathcal{T}, \mathcal{C})$, where $\mathcal{T}$ is a set of real-valued variables, called *time-points,* and $\mathcal{C}$ is a set of binary constraints, called *simple temporal constraints.* Each constraint in $\mathcal{C}$ has the form, $Y - X \leq \delta$, where $X$ and $Y$ are any time-points in $\mathcal{T}$, and $\delta$ is any real number. A *solution* to the STN, $(\mathcal{T}, \mathcal{C})$, is a complete set of assignments to the variables in $\mathcal{T}$ that satisfy all of the constraints in $\mathcal{C}$.

### Conditional Simple Temporal Networks

A Conditional Simple Temporal Problem (CSTP) augments an STN to include observation time-points (or observation nodes) (Tsamardinos, Vidal, and Pollack 2003). Each observation time-point has a proposition associated with it. When the observation time-point is executed, the truth-value of its associated proposition becomes known. In addition, each time-point has a *label* that restricts the scenarios in which that time-point can be executed. For example, a label, $A\neg B$, on a time-point would indicate that that time-point could only be executed in scenarios where the proposition $A$ was true and $B$ was false.

Although not included in the formal definition, the authors made the following *reasonability assumptions* about CSTPs:

(A1) A CSTP should not include any constraint relating time-points whose labels are inconsistent.

(A2) If the label on some time-point $T$ includes a proposition $Q$, then the observation node, $T_Q$, associated with $Q$ must be executed in all cases in which $T$ is executed, and $T_Q$ must be executed *before* $T$ (i.e., $T_Q < T$).[3]

This section defines a Conditional Simple Temporal Network (CSTN), which is the same as a CSTP except that:

- the CSTN definition explicitly incorporates the reasonability assumptions (A1) and (A2) (cf. conditions WD1 and WD2 in Defn. 4, below); and
- each *constraint* in a CSTN has a label associated with it that subsumes the labels of the time-points it constrains (cf. conditions WD1 and WD3 in Defn. 4, below).

Putting labels on the constraints will facilitate the propagation of constraints, as is discussed later on.

**Definition 2** (Label, Label Universe). Given a set $P$ of propositional letters, a *label* is any (possibly empty) conjunction of (positive or negative) literals from $P$. For convenience, the empty label is denoted by $\boxdot$. The *label universe* of $P$, denoted by $P^*$, is the set of all labels whose literals are drawn from $P$.

For example, if $P = \{A, B\}$, then

$$P* = \{\boxdot, A, B, \neg A, \neg B, AB, A\neg B, \neg AB, \neg A\neg B\}.$$

**Definition 3** (Consistent labels, label subsumption).

- Two labels, $\ell_1$ and $\ell_2$, are called *consistent*, denoted by $Con(\ell_1, \ell_2)$, if and only if $\ell_1 \wedge \ell_2$ is satisfiable.
- A label $\ell_1$ *subsumes* a label $\ell_2$, denoted by $Sub(\ell_1, \ell_2)$, if and only if $\models (\ell_1 \Rightarrow \ell_2)$.

To facilitate comparison with the definition of a CSTP, which is not repeated here due to space limitations, the order of arguments in a CSTN is the same as in a CSTP.

**Definition 4** (CSTN). A *Conditional Simple Temporal Network* (CSTN) is a tuple, $\langle \mathcal{T}, \mathcal{C}, L, \mathcal{OT}, \mathcal{O}, P\rangle$, where:

- $\mathcal{T}$ is a finite set of real-valued time-points;
- $P$ is a finite set of propositional letters (or propositions);
- $L : \mathcal{T} \rightarrow P^*$ is a function that assigns a label to each time-point in $\mathcal{T}$;
- $\mathcal{OT} \subseteq \mathcal{T}$ is a (finite) set of observation time-points;
- $\mathcal{O} : P \rightarrow \mathcal{OT}$ is a bijection that associates a unique observation time-point to each propositional letter;

[2] This paper restricts attention to *conditional* connectors. Combi and Posenato (2009) discuss additional kinds of connectors.

[3] Simple temporal constraints do not allow for strict inequalities such as $Y < X$; however, in practice, a constraint such as $Y - X \leq -\epsilon$, for some small $\epsilon > 0$, achieves the desired effect.

- $\mathcal{C}$ is a set of *labeled* simple temporal constraints, each having the form, $(Y - X \leq \delta,\ \ell)$, where $X, Y \in \mathcal{T}$, $\delta$ is a real number, and $\ell \in P^*$;

(WD1) for any labeled constraint, $(Y - X \leq \delta,\ \ell) \in \mathcal{C}$, the label $\ell$ is satisfiable and subsumes both $L(X)$ and $L(Y)$;

(WD2) for each $p \in P$, and each $T \in \mathcal{T}$ for which either $p$ or $\neg p$ appears in $T$'s label,
- $Sub(L(T), L(\mathcal{O}(p))$, and
- $(\mathcal{O}(p) - T \leq -\epsilon, L(T)) \in \mathcal{C}$, for some $\epsilon > 0$; and

(WD3) for each labeled constraint, $(Y - X \leq \delta,\ \ell)$, and for each $p \in P$ for which either $p$ or $\neg p$ appears in $\ell$,
- $Sub(\ell, L(\mathcal{O}(p)))$.

The following definitions will facilitate the proofs of the subsequent lemmas.

**Definition 5** ($\mathcal{C}_{\boxdot}$, $L_{\boxdot}^{\mathcal{T}}$ and $\mathcal{O}_{\emptyset}$)**.**

- If $\mathcal{C}$ is a set of simple temporal constraints, then $\mathcal{C}_{\boxdot}$ is the corresponding set of *labeled* simple temporal constraints, where each constraint is labeled by the empty label, $\boxdot$. In particular, $\mathcal{C}_{\boxdot} = \{(Y - X \leq \delta,\ \boxdot) \mid (Y - X \leq \delta) \in \mathcal{C}\}$
- For any set $\mathcal{T}$ of time-points, $L_{\boxdot}^{\mathcal{T}}$ denotes the labeling function that assigns the empty label to each time-point. Thus, $L_{\boxdot}^{\mathcal{T}}(T) = \boxdot$ for all $T \in \mathcal{T}$. When the context allows, we may write $L_{\boxdot}$ instead of $L_{\boxdot}^{\mathcal{T}}$.
- $\mathcal{O}_{\emptyset}$ denotes the unique function whose domain and range are both empty. Thus, $\mathcal{O}_{\emptyset} : \emptyset \rightarrow \emptyset$.

The following lemmas show that any STN or CSTP can be embedded within a CSTN.

**Lemma 1.** *Let $(\mathcal{T}, \mathcal{C})$ be any STN. Then $\langle \mathcal{T}, \mathcal{C}_{\boxdot}, L_{\boxdot}, \emptyset, \mathcal{O}_{\emptyset}, \emptyset \rangle$ is a CSTN.*

**Proof.** We need only check that the conditions WD1, WD2 and WD3 from the definition of a CSTN are satisfied. WD2 and WD3 are trivially satsified since $P = \emptyset$. As for WD1, each constraint in $\mathcal{C}_{\boxdot}$ has $\boxdot$ as its label, which is satisfiable. Furthermore, $L_{\boxdot}$ assigns the empty label to every node. Thus, the empty label on each constraint trivially subsumes the empty label on the relevant nodes. $\square$

**Lemma 2** (Any CSTP is a CSTN)**.** *Let $\langle V, E, L, OV, \mathcal{O}, P \rangle$ be any CSTP, as defined by Tsamardinos et al. (2003), that satisfies the reasonability assumptions, A1 and A2. Let $\mathcal{S} = \langle V, \mathcal{C}, L, OV, \mathcal{O}, P \rangle$, where:*[4]

$$\mathcal{C} = \bigcup_{(a \leq Y - X \leq b) \in E} \{(a \leq Y - X \leq b, L(X) \wedge L(Y))\}$$

*Then $\mathcal{S}$ is a CSTN.*

**Proof.** Conditions WD1, WD2 and WD3 in the definition of a CSTN (Defn. 4) are satisfied as follows.

(WD1) Each labeled constraint in $\mathcal{C}$ has the form, $(V - U \leq \delta,\ L(U) \wedge L(V))$. Note that $L(U) \wedge L(V)$ subsumes both $L(U)$ and $L(V)$. Furthermore, by assumption A1, $L(U)$ and $L(V)$ must be mutually satisfiable.

(WD2) WD2 is simply a restatement of assumption A2.

(WD3) Each constraint in $\mathcal{C}$ has the form, $(V - U \leq \delta, \ell)$, where $\ell = L(U) \wedge L(V)$. By WD2, $L(U)$ must subsume $L(\mathcal{O}(p))$. But then $\ell$ does too. $\square$

[4] For convenience, we use the expression, $a \leq Y - X \leq b$, to represent the pair of constraints, $Y - X \leq b$ and $X - Y \leq -a$.

### Simple Temporal Networks with Uncertainty

A Simple Temporal Network with Uncertainty (STNU) augments an STN to include a set, $\mathcal{L}$, of contingent links (Morris, Muscettola, and Vidal 2001). Each contingent link has the form, $(A, x, y, C)$, where $A$ and $C$ are time-points, and $0 < x < y < \infty$. $A$ is called the *activation* time-point; $C$ is called the *contingent* time-point. An agent typically activates a contingent link by executing $A$. After doing so, the execution of $C$ is out of the agent's control; however, $C$ is guaranteed to execute such that the temporal difference, $C - A$, is between $x$ and $y$. Contingent links are used to represent actions with uncertain durations; the agent initiates the action, but then merely *observes* its completion in real time.

**Definition 6** (STNU)**.** A Simple Temporal Network with Uncertainty (STNU) is a triple, $(\mathcal{T}, \mathcal{C}, \mathcal{L})$, where:

- $(\mathcal{T}, \mathcal{C})$ is an STN; and
- $\mathcal{L}$ is a set of contingent links, each having the form, $(A, x, y, C)$, where $A$ and $C$ are distinct time-points in $\mathcal{T}$, $0 < x < y < \infty$, and:
  - for each $(A, x, y, C) \in \mathcal{L}$, $\mathcal{C}$ contains the constraints, $(x \leq C - A \leq y)$ (cf. Footnote 4);
  - if $(A_1, x_1, y_1, C_1)$ and $(A_2, x_2, y_2, C_2)$ are distinct contingent links in $\mathcal{L}$, then $C_1$ and $C_2$ are distinct time-points; and
  - the contingent time-point for one contingent link may serve as the activation time-point for another—thus, contingent links may form chains or trees—however, contingent links may not form loops.

As will be seen, the semantics for contingent links is built into the definition of dynamic controllability.

Note that if $(\mathcal{T}, \mathcal{C})$ is an STN, then $(\mathcal{T}, \mathcal{C}, \emptyset)$ is an STNU.

## Conditional STNs with Uncertainty

This section introduces a Conditional STN with Uncertainty (CSTNU) which combines features of CSTNs and STNUs.

**Definition 7** ($\lfloor \mathcal{C} \rfloor$)**.** If $\mathcal{C}$ is a set of labeled constraints of the form, $(Y - X \leq \delta, \ell)$, then $\lfloor \mathcal{C} \rfloor$ is the corresponding set of *unlabeled* constraints:

$$\lfloor \mathcal{C} \rfloor = \{(Y - X \leq \delta) \mid (Y - X \leq \delta, \ell) \in \mathcal{C} \text{ for some } \ell\}.$$

**Definition 8** (CSTNU)**.** A Conditional STN with Uncertainty (CSTNU) is a tuple, $\langle \mathcal{T}, \mathcal{C}, L, \mathcal{OT}, \mathcal{O}, P, \mathcal{L} \rangle$, where:

- $\langle \mathcal{T}, \mathcal{C}, L, \mathcal{OT}, \mathcal{O}, P \rangle$ is a CSTN;
- $(\mathcal{T}, \lfloor \mathcal{C} \rfloor, \mathcal{L})$ is an STNU; and
- for each $(A, x, y, C) \in \mathcal{L}$, $L(A) = L(C)$, and $\mathcal{C}$ contains the labeled constraints, $(x \leq C - A \leq y, L(A))$.[5]

The following lemmas show that any STNU or CSTN can be embedded within a CSTNU.

**Lemma 3.** *If $(\mathcal{T}, \mathcal{C}, \mathcal{L})$ is an STNU, then $\langle \mathcal{T}, \mathcal{C}_{\boxdot}, L_{\boxdot}, \emptyset, \mathcal{O}_{\emptyset}, \emptyset, \mathcal{L} \rangle$ is a CSTNU.*

[5] $(x \leq C - A \leq y, L(A))$ is shorthand for the pair of labeled constraints, $(A - C \leq -x,\ L(A))$ and $(C - A \leq y, L(A))$.

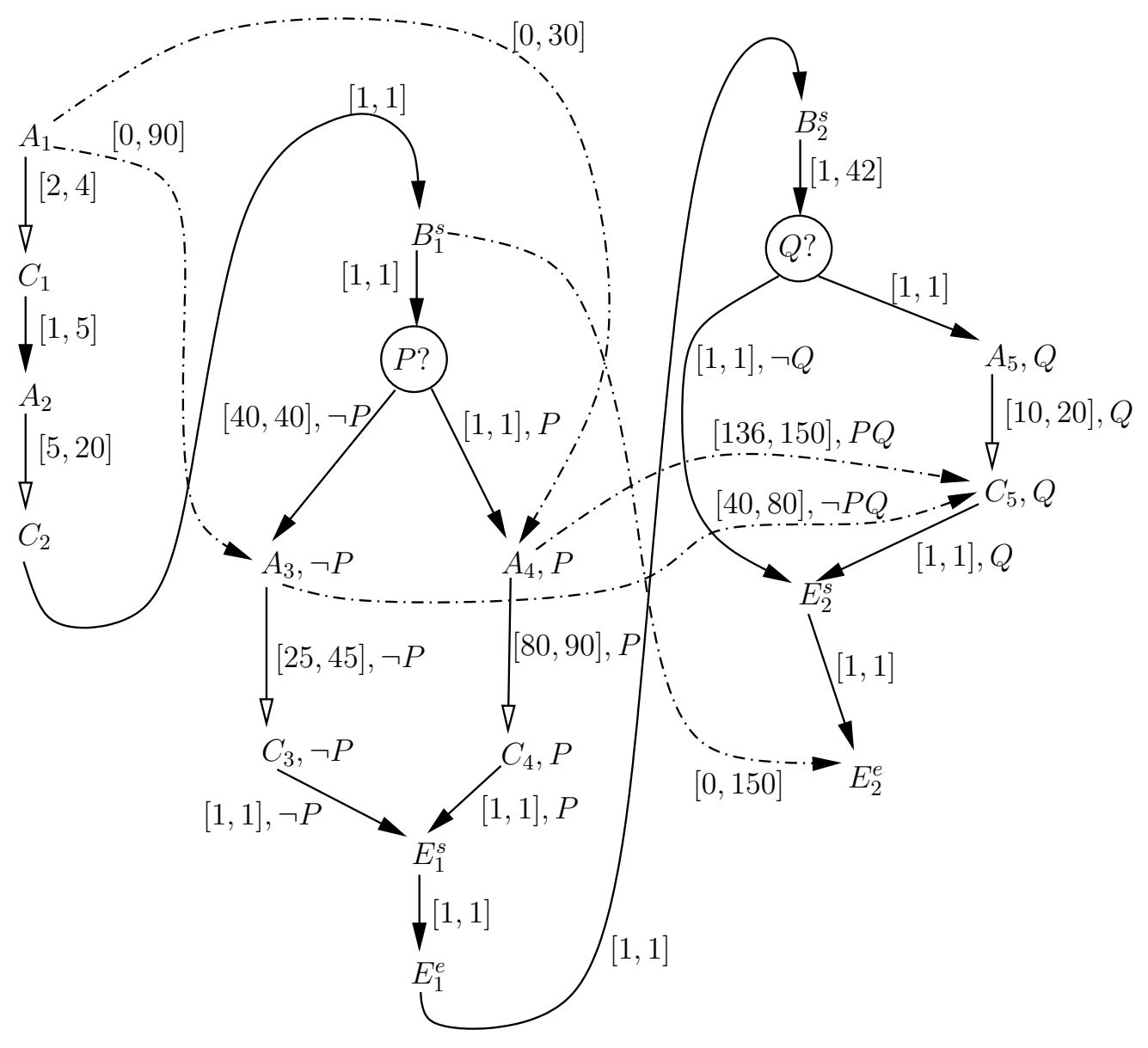

*Figure 2:* The CSTNU for the workflow in Fig. 1

**Proof.** Let $(\mathcal{T}, \mathcal{C}, \mathcal{L})$ be any STNU. Then $(\mathcal{T}, \mathcal{C})$ is an STN. By Lemma 1, $\langle \mathcal{T}, \mathcal{C}_\boxdot, L_\boxdot, \emptyset, \mathcal{O}_\emptyset, \emptyset \rangle$ is a CSTN. In addition, since $\lfloor \mathcal{C}_\boxdot \rfloor$ is necessarily the same as $\mathcal{C}$, $(\mathcal{T}, \lfloor \mathcal{C}_\boxdot \rfloor, \mathcal{L})$ must be an STNU. Finally, for each $(A, x, y, C) \in \mathcal{L}$, $\mathcal{C}$ contains the constraints, $(x \leq C - A \leq y)$, which implies that $\mathcal{C}_\boxdot$ contains the labeled constraints, $(x \leq C - A \leq y, \boxdot)$. Since $L_\boxdot$ assigns the empty label to each node, the last condition of Defn. 8 is satisfied. $\Box$

**Lemma 4.** *If* $\langle \mathcal{T}, \mathcal{C}, L, \mathcal{OT}, \mathcal{O}, P \rangle$ *is an CSTN, then* $\langle \mathcal{T}, \mathcal{C}, L, \mathcal{OT}, \mathcal{O}, P, \emptyset \rangle$ *is a CSTNU.*

**Proof.** Let $\langle \mathcal{T}, \mathcal{C}, L, \mathcal{OT}, \mathcal{O}, P \rangle$ be any CSTN. Then $(\mathcal{T}, \lfloor \mathcal{C} \rfloor)$ is an STN, whence $(\mathcal{T}, \lfloor \mathcal{C} \rfloor, \emptyset)$ is an STNU. Since $\mathcal{L}$ is empty, the last condition of Defn. 8 is satisfied. $\Box$

### The CSTNU Underlying a Worklow

Recall the sample workflow from Fig. 1. This workflow has an underlying CSTNU that is derived by

- replacing each task with a corresponding contingent link;
- replacing each split connector by a pair of (starting and ending) time-points, linked by a duration constraint, and where the ending time-point is an observation node for a proposition whose possible values correspond to the different branching decisions; and
- replacing each join connector by a pair of time-points, linked by a duration constraint.

Fig. 2 shows the CSTNU underlying the workflow from Fig. 1. In this CSTNU, each contingent link from $A_i$ to $C_i$ corresponds to the task $T_i$ from the workflow; and observation nodes are circled. Note that the branch containing task $T_4$ is labeled by $P$, whereas the alternative branch containing task $T_3$ is labeled by $\neg P$. Similarly, the branch containing task $T_5$ is labeled by $Q$, and the alternative branch is labeled by $\neg Q$. Note, too, that labels on edges subsume the labels on the time-points they connect. Dashed edges are kept dashed to facilitate comparison with the workflow in Fig. 1.

## Dynamic Controllability

This section combines the semantics of CSTNs and STNUs to generate a definition for the dynamic controllability of a CSTNU. Because the semantics for the corresponding notions involve similar definitions, in some cases the various terms, such as *history* or *dynamic* will be given prefixes or superscripts to indicate the kinds of networks or situations/scenarios they apply to. In addition, we use the term, *history*, instead of *pre-history*, for convenience.

### Dynamic Consistency of CSTNs

A CSTP is called *dynamically consistent* if there exists a strategy for executing its time-points that guarantees the satisfaction of all relevant constraints no matter how the truth values of the various observations turn out (Tsamardinos, Vidal, and Pollack 2003). The strategy is dynamic in that its execution decisions can react to past observations, but not those in the future. This section defines the dynamic consistency of a CSTN in an equivalent way; however, for convenience, there are some superficial differences in notation and organization. Afterward, we provide a second characterization of the *dynamic* property that will be useful later on.

**Definition 9** (Scenario/Interpretation Function)**.** A *scenario* (or *interpretation function*) over a set $P$ of propositional letters is a function, $s : P \rightarrow \{true, false\}$, that assigns a truth value to each letter in $P$.[6] As is standard practice in propositional logic, any interpretation function can be extended to provide the truth value for every possible formula involving the letters in $P$. Thus, any interpretation function, $s$, can provide the truth value of each label involving letters in $P$. For any label, $\ell$, the truth value of $\ell$ in the scenario, $s$, is denoted by $s(\ell)$. Let $\mathcal{I}_P$ (or simply $\mathcal{I}$) denote the set of all interpretation functions (or complete execution scenarios) over $P$.

**Definition 10** (Schedule)**.** A *schedule* for a set of time-points $\mathcal{T}$ is a mapping, $\psi : \mathcal{T} \rightarrow \mathbb{R}$ that assigns a real number to each time-point in $\mathcal{T}$. The set of all schedules *for any subset of* $\mathcal{T}$ is denoted by $\Psi_\mathcal{T}$ (or $\Psi$ if the context allows).

Below, the projection of a CSTN, $\mathcal{S}$, onto a scenario, $s$, is defined to be the STN that contains all of the time-points and constraints from $\mathcal{S}$ whose labels are true under $s$ (i.e., the time-points that must be executed under $s$, and the constraints that must be satisfied under $s$).

**Definition 11** (Scenario Projection for a CSTN)**.** Let $\mathcal{S} = \langle \mathcal{T}, \mathcal{C}, L, \mathcal{OT}, \mathcal{O}, P \rangle$ be any CSTN, and $s$ any interpretation function (i.e., complete scenario) for the letters in $P$. The *projection* of $\mathcal{S}$ onto the scenario $s$—denoted by $scPrj(\mathcal{S}, s)$—is the STN, $(\mathcal{T}_s^+, \mathcal{C}_s^+)$, where:

- $\mathcal{T}_s^+ = \{T \in \mathcal{T} : s(L(T)) = true\}$; and
- $\mathcal{C}_s^+ = \{(Y - X \leq \delta) \mid$ for some $\ell$, $(Y - X \leq \delta, \ell) \in \mathcal{C}$ and $s(\ell) = true\}$

[6] Unlike the prior work on CSTPs, we restrict attention to *complete* scenarios because the subsequent definition of a *history* requires a scenario to entail the outcome of *all* past observations.

Recall that condition WD1 from the definition of a CSTN stipulates that the label on any constraint must subsume the labels on the time-points it connects. Thus, for any constraint in $\mathcal{C}_s^+$, the time-points it connects must belong to the set $\mathcal{T}_s^+$.

**Definition 12** (Execution Strategy for a CSTN)**.** Let $\mathcal{S} = \langle \mathcal{T}, \mathcal{C}, L, \mathcal{OT}, \mathcal{O}, P\rangle$ be any CSTN. An *execution strategy* for $\mathcal{S}$ is a mapping, $\sigma : \mathcal{I} \rightarrow \Psi_{\mathcal{T}}$, such that for each scenario, $s \in \mathcal{I}$, the domain of $\sigma(s)$ is $\mathcal{T}_s^+$ (cf. Defn. 11). If, in addition, for each scenario, $s$, the schedule $\sigma(s)$ is a solution to the scenario projection, $scPrj(\mathcal{S}, s)$, then $\sigma$ is called *viable.* In any case, the execution time for the time-point $X$ in the schedule $\sigma(s)$ is denoted by $[\sigma(s)]_X$.

Below, the *history* of a time-point, $X$, relative to a scenario, $s$, and strategy, $\sigma$, is defined to be the set of observations made before the time at which $X$ is executed according to the schedule, $\sigma(s)$ (i.e., before the time $[\sigma(s)]_X$).[7]

**Definition 13** (Scenario history for a CSTN)**.** Let $\mathcal{S} = \langle \mathcal{T}, \mathcal{C}, L, \mathcal{OT}, \mathcal{O}, P\rangle$ be any CSTN, $s$ any scenario, $\sigma$ any execution strategy for $\mathcal{S}$, and $X$ any time-point in $\mathcal{T}_s^+$ (cf. Defn. 11). The *history* of $X$ in the scenario $s$, for the strategy $\sigma$—denoted by $scHst(X, s, \sigma)$—is given by:

$$scHst(X, s, \sigma) \;=\; \{(p, s(p)) \mid \mathcal{O}(p) \in \mathcal{T}_s^+, \text{ and } [\sigma(s)]_{\mathcal{O}(p)} < [\sigma(s)]_X\}$$

Note that any scenario history determines a corresponding label whose (positive or negative) literals are in a one-to-one correspondence with the observations, $(p, s(p))$, in the history. Thus, we may sometimes (e.g., in the next definition) treat a scenario history as though it were a label.

Below, an execution strategy is called *dynamic* if the schedules it generates always assign the same execution time to any time-point $X$ in scenarios that cannot be distinguished prior to that time.[8]

**Definition 14** (Dynamic Execution Strategy for a CSTN)**.** An execution strategy, $\sigma$, for a CSTN is called *dynamic* if for all scenarios, $s_1$ and $s_2$, and any time-point $X$:

if $Con(s_1, scHst(X, s_2, \sigma))$, then $[\sigma(s_1)]_X = [\sigma(s_2)]_X$.

**Definition 15** (Dynamic Consistency for a CSTN)**.** A CSTN is called *dynamically consistent* if there exists an execution strategy for it that is both viable and dynamic.

The following definitions and lemma provide an equivalent, alternative characterization of a dynamic execution strategy for a CSTN. First, a scenario history relative to a numerical time—not a time-point variable—is defined.

**Definition 16** (Scenario History* for a CSTN)**.** Let $\mathcal{S} = \langle \mathcal{T}, \mathcal{C}, L, \mathcal{OT}, \mathcal{O}, P\rangle$ be any CSTN, $s$ any scenario, $\sigma$ any execution strategy for $\mathcal{S}$, and $t$ any real number. The *history** of $t$ in the scenario $s$, for the strategy $\sigma$—denoted by $scHst^*(t, s, \sigma)$—is the set of all observations made before time $t$ according to the schedule, $\sigma(s)$. In particular:

$$scHst^*(t, s, \sigma) = \{(p, s(p)) \mid \mathcal{O}(p) \in \mathcal{T}_s^+ \text{ and } [\sigma(s)]_{\mathcal{O}(p)} < t\}$$

Note that for all time-points $X$, scenarios $s$, and strategies $\sigma$, $scHst(X, s, \sigma) = scHst^*([\sigma(s)]_X, s, \sigma)$.

**Definition 17** (Dynamic* Execution Strategy for a CSTN)**.** An execution strategy, $\sigma$, for an CSTN is called *dynamic** if for any scenarios, $s_1$ and $s_2$, and any time-point, $X$:

if $scHst^*([\sigma(s_1)]_X, s_1, \sigma) = scHst^*([\sigma(s_1)]_X, s_2, \sigma)$, then $[\sigma(s_1)]_X = [\sigma(s_2)]_X$.

Notice that in this definition, the two histories, one relative to $s_1$, the other to $s_2$, are taken with respect to the *same* (numeric) time, $[\sigma(s_1)]_X$. If the strategy $\sigma$ yields schedules for $s_1$ and $s_2$ that have identical histories prior to that one time, then those schedules must assign the same value to $X$.

**Lemma 5.** *An execution strategy $\sigma$ for a CSTN is dynamic if and only if it is dynamic*$^*$.

**Proof.**

($\Rightarrow$) Suppose $\sigma$ is a dynamic execution strategy for some CSTN. Let $s_1$ and $s_2$ be any scenarios, and $X$ any time-point such that $scHst^*(t_1, s_1, \sigma) = scHst^*(t_1, s_2, \sigma)$, where $t_1 = [\sigma(s_1)]_X$. Now $s_2$ must be consistent with $scHst^*(t_1, s_2, \sigma)$, since the observations contained in that history are a subset of the observations in $s_2$. Thus, $s_2$ is consistent with $scHst^*(t_1, s_1, \sigma)$, which equals $scHst(X, s_1, \sigma)$. Thus, since $\sigma$ is dynamic, we must have that $[\sigma(s_1)]_X = [\sigma(s_2)]_X$. Thus, $\sigma$ is dynamic*.

($\Leftarrow$) Suppose $\sigma$ is a dynamic* execution strategy for some CSTN. Let $s_1$ and $s_2$ be any scenarios, and $X$ any time-point such that $Con(s_1, scHist(X, s_2, \sigma))$. Suppose that $[\sigma(s_1)]_X \neq [\sigma(s_2)]_X$. Let $t \in \mathbb{R}$ be the first time at which the schedules $\sigma(s_1)$ and $\sigma(s_2)$ diverge. Then, $t \leq \min\{[\sigma(s_1)]_X, [\sigma(s_2)]_X\}$; and there must be some time-point $Y$ that is executed at time $t$ in one scenario, and at some later time in the other scenario.

By construction, $t \leq [\sigma(s_2)]_X$. Thus, $scHst^*(t, s_2, \sigma) \subseteq scHst^*([\sigma(s_2)]_X, s_2, \sigma) = scHst(X, s_2, \sigma)$. Thus, since $Con(s_1, scHst(X, s_2, \sigma))$, it follows that $Con(s_1, scHst^*(t, s_2, \sigma))$. And since $s_1$ is a complete scenario, the observations in $scHst^*(t, s_2, \sigma)$ must be a subset of the "observations" in $s_1$. And since, by construction, the schedules, $\sigma(s_1)$ and $\sigma(s_2)$, are identical prior to time $t$, it follows that the observations in the two histories, $scHst^*(t, s_1, \sigma)$ and $scHst^*(t, s_2, \sigma)$, involve the same sets of observation time-points with identical outcomes (i.e., truth values). Thus, $scHst^*(t, s_1, \sigma) = scHst^*(t, s_2, \sigma)$, whence the property of $\sigma$ being dynamic* implies that $[\sigma(s_1)]_Y = [\sigma(s_2)]_Y$, contradicting the choice of $Y$. Thus, it must be that the schedules, $\sigma(s_1)$ and $\sigma(s_2)$ diverge, if at all, *after* the execution of $X$, in which case, $[\sigma(s_1)]_X = [\sigma(s_2)]_X$. Thus, $\sigma$ is dynamic. $\square$

**Dynamic Controllability of STNUs**

Morris et al. (2001) call an STNU *dynamically controllable* if there exists a strategy for executing its time-points that guarantees the satisfaction of all constraints in the network no matter how the durations of the contingent links turn out. The strategy is dynamic in that its execution decisions can react to observations of contingent links that have already completed, but not to those that have yet to complete.

[7]Tsamardinos et al. (2003) define (pre)histories for arbitrary schedules, whereas here we restrict attention to schedules of the form, $\sigma(s)$, where $\sigma$ is an execution strategy and $s$ is a scenario.

[8]Tsamardinos et al. (2003) include a disjunctive condition, $Con(s_1, scHst(X, s_2, \sigma)) \vee Con(s_2, scHist(X, s_1, \sigma))$. However, since $s_1$ and $s_2$ play symmetric roles in the two disjuncts, and since $s_1$ and $s_2$ are both universally quantified (cf. Defn. 14), it suffices to include just one of the disjuncts.

This section presents a sequence of definitions that culminate in the definition of the dynamic controllability of an STNU. Most of the definitions are from Morris et al. (2001), albeit with some slight differences in notation, but *history*$^*$ and *dynamic*$^*$ are from Hunsberger (2009). Parallels between the definitions in this section and those from the preceding section are highlighted along the way.

Analogous to a scenario for a CSTN, which specifies the truth value for each proposition, a *situation* for an STNU specifies fixed durations for all of the contingent links.

**Definition 18** (Situations). Let $\mathcal{S}$ be an STNU having the $k$ contingent links, $(A_1, x_1, y_1, C_1), \ldots, (A_k, x_k, y_k, C_k)$, with respective duration ranges, $[x_1, y_1], \ldots, [x_k, y_k]$. Then $\Omega_{\mathcal{S}} = [x_1, y_1] \times \ldots \times [x_k, y_k]$ is called the *space of situations* for $\mathcal{S}$. Any $\omega = (d_1, \ldots, d_k) \in \Omega_{\mathcal{S}}$ is called a *situation.* When context allows, we may write $\Omega$ instead of $\Omega_{\mathcal{S}}$.

Schedules for STNUs are defined the same way as for CSTNs, except that the domain for each schedule must be the entire set of time-points, $\mathcal{T}$.

The projection of a CSTN onto a scenario yields an STN by fixing the truth value of each propositional letter and restricting attention to those time-points and constraints whose labels are true according to that scenario. Analogously, the projection of an STNU onto a situation yields an STN by fixing the duration of each contingent link.

**Definition 19** (Situation Projection for an STNU). Suppose $\mathcal{S} = (\mathcal{T}, \mathcal{C}, \mathcal{L})$ is an STNU and $\omega = (d_1, \ldots, d_k)$ is a situation. The *projection* of $\mathcal{S}$ onto the situation $\omega$—denoted by $sitPrj(\mathcal{S}, \omega)$—is the STN, $(\mathcal{T}, \mathcal{C}'')$, where:

$$\mathcal{C}'' = \mathcal{C} \cup \{(d_i \leq C_i - A_i \leq d_i) \mid 1 \leq i \leq k\}.$$

**Definition 20** (Execution Strategy for an STNU). Let $\mathcal{S} = (\mathcal{T}, \mathcal{C}, \mathcal{L})$ be an STNU. An *execution strategy* for $\mathcal{S}$ is a mapping, $\sigma : \Omega \to \Psi$, such that for each situation, $\omega \in \Omega$, $\sigma(\omega)$ is a (complete) schedule for the time-points in $\mathcal{T}$. If, in addition, for each situation, $\omega$, the schedule $\sigma(\omega)$ is a solution for the situation projection, $sitPrj(\mathcal{S}, \omega)$, then $\sigma$ is called *viable.* In any case, the execution time for any time-point $X$ in the schedule, $\sigma(\omega)$, is denoted by $[\sigma(\omega)]_X$.

Analogous to a scenario history$^*$ for a CSTN, a situation history$^*$ for an STNU specifies the durations of all contingent links that have finished executing prior to a (numeric) time $t$ in a schedule $\sigma(\omega)$.

**Definition 21** (Situation History$^*$ for an STNU). Let $\mathcal{S} = (\mathcal{T}, \mathcal{C}, \mathcal{L})$ be any STNU, $\sigma$ any execution strategy for $\mathcal{S}$, $\omega$ any situation, and $t$ any real number. The *history*$^*$ of $t$ in the situation $\omega$, for the strategy $\sigma$—denoted by $sitHst(t, \omega, \sigma)$—is the set:

$$sitHst(t, \omega, \sigma) = \{(A, C, [\sigma(\omega)]_C - [\sigma(\omega)]_A) \mid \exists x, y \text{ s.t. } (A, x, y, C) \in \mathcal{L} \text{ and } [\sigma(\omega)]_C < t\}$$

The definition of the *dynamic*$^*$ property for an execution strategy for an STNU parallels that of the *dynamic*$^*$ property for an execution strategy for a CSTN.

**Definition 22** (Dynamic$^*$ Execution Strategy for an STNU). An execution strategy, $\sigma$, for an STNU is called *dynamic*$^*$ if for any situations, $\omega_1$ and $\omega_2$, and any *non-contingent* time-point $X$:

if $sitHst([\sigma(\omega_1)]_X, \omega_1, \sigma) = sitHst([\sigma(\omega_1)]_X, \omega_2, \sigma)$, then $[\sigma(\omega_1)]_X = [\sigma(\omega_2)]_X$.

**Definition 23** (Dynamic Controllability for an STNU). An STNU $\mathcal{S}$ is called *dynamically controllable* if there exists an execution strategy for $\mathcal{S}$ that is both viable and dynamic$^*$.

## Dynamic Controllability of CSTNUs

This section extends the notions of the dynamic consistency of a CSTN and the dynamic controllability of an STNU to generate a (novel) definition of the dynamic controllability of a CSTNU. To wit, a sequence of definitions is presented that parallels those of the preceding sections.

A *drama* is a scenario/situation pair that specifies fixed truth values for all of the propositional letters and fixed durations for all of the contingent links.

**Definition 24** (Drama). Given a CSTNU $\mathcal{S}$, a *drama* is any pair $(s, \omega)$, where $s$ is a scenario, and $\omega$ is a situation. The set of all dramas (for $\mathcal{S}$) is $\mathcal{I} \times \Omega$.

Next, the *projection* of a CSTNU onto a drama, $(s, \omega)$, is defined. The projection restricts attention to time-points and constraints whose labels are true under the scenario $s$, while also including constraints that force the contingent links to take on the durations specified in the situation $\omega$.

**Definition 25** (Drama Projection for a CSTNU). Suppose $\mathcal{S} = \langle \mathcal{T}, \mathcal{C}, L, \mathcal{OT}, \mathcal{O}, P, \mathcal{L} \rangle$ is a CSTNU and $(s, \omega)$ is a drama for $\mathcal{S}$, where $\omega = (d_1, \ldots, d_k)$. The *projection* of $\mathcal{S}$ onto the drama $(s, \omega)$—denoted by $drPrj(\mathcal{S}, s, \omega)$—is the STN, $(\mathcal{T}_s^+, \mathcal{C}_1 \cup \mathcal{C}_0)$, where:

- $\mathcal{T}_s^+ = \{T \in \mathcal{T} : s(L(T)) = true\}$
- $\mathcal{C}_1 = \{(Y - X \leq \delta) \mid \text{ for some } \ell, (Y - X \leq \delta, \ell) \in \mathcal{C}, \text{ and } s(\ell) = true\}$
- $\mathcal{C}_0 = \{(d_i \leq C_i - A_i \leq d_i) \mid (A_i, x_i, y_i, C_i) \in \mathcal{L} \text{ and } A_i, C_i \in \mathcal{T}_s^+\}$

**Definition 26** (Execution Strategy for a CSTNU). Let $\mathcal{S} = \langle \mathcal{T}, P, L, \mathcal{OT}, \mathcal{O}, \mathcal{C}, \mathcal{L} \rangle$ be a CSTNU. An *execution strategy* for $\mathcal{S}$ is a mapping, $\sigma : (\mathcal{I} \times \Omega) \to \Psi_{\mathcal{T}}$, such that for each drama, $(s, \omega)$, the domain of $\sigma(s, \omega)$ is $\mathcal{T}_s^+$. $\sigma$ is called *viable* if for each drama, $(s, \omega)$, the schedule $\sigma(s, \omega)$ is a solution to the projection, $drPrj(s, \omega)$. For any time-point $X$ and drama $(s, \omega)$, the execution time of $X$ in the schedule, $\sigma(s, \omega)$, is denoted by $[\sigma(s, \omega)]_X$.

The following definition combines the definitions of history$^*$ relative to a numeric time for CSTNs and STNUs.

**Definition 27** (Drama History$^*$ for a CSTNU). Let $\mathcal{S} = \langle \mathcal{T}, P, L, \mathcal{OT}, \mathcal{O}, \mathcal{C}, \mathcal{L} \rangle$ be a CSTNU. Let $\sigma$ be an execution strategy for $\mathcal{S}$, $(s, \omega)$ some drama, and $t$ some real number. Then the *history*$^*$ of $t$ for the drama $(s, \omega)$ and strategy $\sigma$—denoted by $drHst(t, s, \omega, \sigma)$—is the pair $(\mathcal{H}_s, \mathcal{H}_\omega)$ where:

- $\mathcal{H}_s = \{(p, s(p)) \mid \mathcal{O}(p) \in \mathcal{T}_s^+ \text{ and } [\sigma(s, \omega)]_{\mathcal{O}(p)} < t\}$; and

- $\mathcal{H}_\omega = \{(A, C, [\sigma(s,\omega)]_C - [\sigma(s,\omega)]_A) \mid A, C \in \mathcal{T}_s^+,$
  $\exists x, y$ s.t. $(A, x, y, C, ) \in \mathcal{L},\ [\sigma(s,\omega)]_C < t\}$.

Note that $\mathcal{H}_s$ specifies the truth values of all propositions that are observed prior to $t$ in the schedule $\sigma(s,\omega)$; and $\mathcal{H}_\omega$ specifies the durations of all contingent links that finish executing prior to $t$ in that schedule.

**Definition 28** (Dynamic$^*$ Execution Strategy for a CSTNU)**.** An execution strategy, $\sigma$, for a CSTNU is called *dynamic*$^*$ if for every pair of dramas, $(s_1, \omega_1)$ and $(s_2, \omega_2)$, and every *non-contingent* time-point $X \in \mathcal{T}_{s_1}^+ \cap \mathcal{T}_{s_2}^+$:

if $drHst(t, s_1, \omega_1, \sigma) = drHst(t, s_2, \omega_2, \sigma)$,
where $t = [\sigma(s_1, \omega_1)]_X$,
then $[\sigma(s_1, \omega_1)]_X = [\sigma(s_2, \omega_2)]_X$.

**Definition 29** (Dynamic Controllability for a CSTNU)**.** A CSTNU, $\mathcal{S}$, is *dynamically controllable* if there exists an execution strategy for $\mathcal{S}$ that is both viable and dynamic$^*$.

The following lemmas show that the above definition properly generalizes the dynamic consistency of a CSTN and the dynamic controllability of an STNU.

**Lemma 6.** *Let $\mathcal{S} = \langle \mathcal{T}, \mathcal{C}, L, \mathcal{OT}, \mathcal{O}, P \rangle$ be any CSTN. Then $\mathcal{S}$ is dynamically consistent if and only if the CSTNU, $\mathcal{S}_u = \langle \mathcal{T}, \mathcal{C}, L, \mathcal{OT}, \mathcal{O}, P, \emptyset \rangle$, is dynamically controllable.*

**Proof.** Let $\mathcal{S} = \langle \mathcal{T}, \mathcal{C}, L, \mathcal{OT}, \mathcal{O}, P \rangle$ be any dynamically consistent CSTN. Then $\mathcal{S}$ has an execution strategy, $\sigma : \mathcal{I} \rightarrow \Psi_\mathcal{T}$, that is both viable and dynamic. By Lemma 5, $\sigma$ is also dynamic$^*$. In addition, since $\mathcal{S}$ is a CSTN, Lemma 4 ensures that $\mathcal{S}_u = \langle \mathcal{T}, \mathcal{C}, L, \mathcal{OT}, \mathcal{O}, P, \emptyset \rangle$ is a CSTNU. We must show that $\mathcal{S}_u$ has an execution strategy, $\sigma_u : (\mathcal{I} \times \Omega) \rightarrow \Psi_\mathcal{T}$, that is both viable and dynamic$^*$. Note that since $\mathcal{S}_u$ has no contingent links, $\Omega$ contains exactly one situation—the null situation—which we shall denote by $\omega_\emptyset$.

Define $\sigma_u$ as follows. For any drama, $(s, \omega_\emptyset)$, let $\sigma_u(s, \omega_\emptyset) = \sigma(s)$. Note that $\sigma_u$ is an execution strategy for $\mathcal{S}_u$, since the domain of $\sigma(s)$ is guaranteed to be $\mathcal{T}_s^+$.

Since $\sigma$ is viable, for any scenario $s$, the schedule $\sigma(s)$ is a solution to the scenario projection, $scPrj(\mathcal{S}, s)$. However, for any $s$, the schedules, $\sigma(s)$ and $\sigma(s, \omega_\emptyset)$ are defined to be the same. Furthermore, since $\mathcal{S}_u$ has no contingent links, it follows that for any $s$, the drama projection, $drPrj(\mathcal{S}_u, s, \omega_\emptyset)$, is the same STN as $scPrj(\mathcal{S}, s)$ (cf. Defns. 11 and 25). Thus, for any $s$, $\sigma(s, \omega_\emptyset)$ is necessarily a solution to $drPrj(\mathcal{S}_u, s, \omega_\emptyset)$, whence $\sigma_u$ is viable.

To show that $\sigma_u$ is dynamic$^*$, suppose $(s_1, \omega_\emptyset)$ and $(s_2, \omega_\emptyset)$ are any dramas in $\mathcal{I} \times \Omega$, $X$ is a non-contingent time-point in $\mathcal{T}_{s_1}^+ \cap \mathcal{T}_{s_2}^+$, $t = [\sigma_u(s_1, \omega_\emptyset)]_X$, and $drHst^*(t, s_1, \omega_\emptyset, \sigma_u) = drHst^*(t, s_2, \omega_\emptyset, \sigma_u)$. Note that $t = [\sigma_u(s_1, \omega_\emptyset)]_X = [\sigma(s_1)]_X$. Furthermore, since there are no contingent links, $drHst^*(t, s_1, \omega_\emptyset, \sigma_u) = drHst^*(t, s_2, \omega_\emptyset, \sigma_u)$ if and only if $scHst^*(t, s_1, \sigma) = scHst^*(t, s_2, \sigma)$ (cf. Defns. 16 and 27). But then $\sigma$ being dynamic$^*$ ensures that $[\sigma(s_1)]_X = [\sigma(s_2)]_X$ (cf. Defn. 17), and hence $[\sigma_u(s_1, \omega_\emptyset)]_X = [\sigma_u(s_2, \omega_\emptyset)]_X$. $\square$

**Lemma 7.** *Let $\mathcal{S} = (\mathcal{T}, \mathcal{C}, \mathcal{L})$ be any STNU. Then $\mathcal{S}$ is dynamically controllable if and only if the CSTNU, $\langle \mathcal{T}, \mathcal{C}_\boxdot, L_\boxdot, \emptyset, \mathcal{O}_\emptyset, \emptyset, \mathcal{L} \rangle$, is dynamically controllable.*

**Proof.** The proof is omitted for space reasons. It has the same general structure as the proof of Lemma 6.

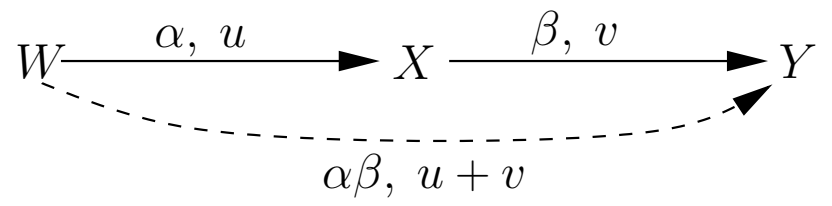

*Figure 3:* Basic constraint propagation in a CSTNU

## Toward a DC-Checking Algorithm for CSTNUs

This section addresses the problem of finding an algorithm for determining the dynamic controllability of arbitrary CSTNUs. Given that CSTNUs combine the features of CSTPs and STNUs, one approach would be to combine existing algorithms for determining the dynamic consistency of CSTPs and the dynamic controllability of STNUs. However, those algorithms employ very different techniques. For example, in the CSTP algorithm, Tsamardinos et al. (2003) first derive a related Disjunctive Temporal Problem (DTP), and then solve it using a dedicated DTP solver that is optimized by a variety of constraint-satisfaction heuristics. In contrast, the fastest algorithm for determining whether arbitrary STNUs are dynamically controllable is the $O(N^4)$-time algorithm developed by Morris (2006), which is a constraint-propagation algorithm that focuses on the *reducing away of lower-case edges* in an STNU graph.

Another problem is that the CSTP algorithm uses exponential space and time. Conrad and colleagues (Conrad 2010; Conrad and Williams 2011) developed the Drake system for propagating *labeled* constraints in temporal networks with choice.[9] The aim was to reduce the space required to generate dispatchable plans, while accepting slight increases in the time requirements. Although their choice nodes are dramatically different from the observation nodes in a CSTNU—because choice nodes are *controlled* by the agent—their use of *labeled value sets* in constraint propagation inspired our use of labels on the edges of a CSTNU.

### Constraint Propagation for CSTNUs

Consider the propagation of labeled constraints illustrated in Fig. 3. Any dynamic execution strategy that observes the labeled constraints from $W$ to $X$, and from $X$ to $Y$, must also observe the derived constraint from $W$ to $Y$. Notice that the label on the derived constraint is the conjunction of the labels on the original constraints. The proof that this propagation rule is sound is omitted, due to space limitations.

### Label Modification in a CSTNU

Morris et al. (2001) showed that the presence of contingent links in an STNU requires new kinds of constraint propagation when checking dynamic controllability. Those kinds of rules will also be needed for a CSTNU. However, in addition, the presence of observation nodes requires new kinds of propagation rules. One such rule is presented below.

Consider the CSTNU fragment in Fig. 4, where $0 \leq w$, $v \leq w$, $\alpha, \beta$ and $\gamma$ are labels that do not share any propositional letters, and $p$ is a propositional letter that does not appear in $\alpha$, $\beta$ or $\gamma$. The time-point, $p?$, is the observation time-point for $p$. Thus, when $p?$ executes, the value of $p$ becomes known. The arrow from $p?$ to $X$ represents the labeled constraint, $(X - p? \leq -w,\ \alpha\beta)$. Thus, in scenarios where $\alpha\beta$ is true, $X + w \leq p?$ must hold. Thus, in those scenarios, $X$ must be executed before $p$ is observed. The arrow from $X$ to $Y$ represents the constraint, $(Y - X \leq v,\ \beta\gamma p)$. Thus, in scenarios where $\beta\gamma p$ is true, $Y \leq X + v$ must hold—in which case, $Y$ must execute before the value of $p$ is known.

[9] In the earlier paper, they incorporated contingent links and presented a preliminary extension of their dispatchability algorithm.

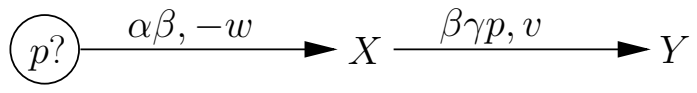

*Figure 4:* The context for label modification in a CSTNU

**Lemma 8** (Label Modification Rule)**.** *If $\sigma$ is a dynamic execution strategy that satisfies the labeled constraints in Fig. 4 in all scenarios where their labels are true, then $\sigma$ must also satisfy the labeled constraint, $(Y - X \leq v,\ \alpha\beta\gamma)$, in all scenarios where $\alpha\beta\gamma$ is true. Moreover, the original labeled constraint, $(Y - X \leq v,\ \beta\gamma p)$, can be replaced by the pair of labeled constraints, $(Y - X \leq v,\ \alpha\beta\gamma)$ and $(Y - X \leq v,\ (\neg\alpha)\beta\gamma p)$.*

**Proof.** Let $\sigma$ be as in the statement of the lemma. However, suppose that there is some drama, $(s, \omega)$, such that: (1) the label $\alpha\beta\gamma$ is true in scenario $s$; but (2) the schedule, $\sigma(s, \omega)$, does *not* satisfy the constraint, $(Y - X \leq v)$. Let $s_2$ be the same scenario as $s$, except that the value of $p$ is flipped. Now, by construction, in one of the scenarios, $s$ or $s_2$, the label, $\alpha\beta\gamma p$, is true. Let $\hat{s}$ be that scenario, and $\sigma(\hat{s}, \omega)$ the corresponding schedule. By construction, that schedule satisfies both of the labeled constraints from Fig. 4, since their labels are true in $\hat{s}$. Thus,

$$\begin{aligned}[\sigma(\hat{s},\omega)]_Y &\leq [\sigma(\hat{s},\omega)]_X + v\,, && \text{since } Y - X \leq v\\ &\leq [\sigma(\hat{s},\omega)]_X + w, && \text{since } v \leq w\\ &\leq p? \qquad\qquad\quad , && \text{since } X - p? \leq -w\end{aligned}$$

Let $\tilde{s}$ be the scenario that is the same as $\hat{s}$, except that the value of $p$ is flipped. Let $t$ be the first time at which the schedules, $\sigma(\hat{s}, \omega)$ and $\sigma(\tilde{s}, \omega)$, differ. Thus, there must be some time-point $T$ that is executed in one of the schedules at time $t$, and in the other at some time later than $t$. But in that case, the corresponding histories at time $t$ must be different. But the only possible difference must involve the value of the proposition $p$, since all other propositions and contingent durations are identical in the dramas, $(\hat{s}, \omega)$ and $(\tilde{s}, \omega)$. Thus, $p?$ must be executed before time $t$. Since $t$ is the time of first difference in the schedules, it follows that $p?$ is executed at the same time in each of these schedules. Furthermore, since $X$ and $Y$ are both executed before $p?$ in $\sigma(\hat{s}, \omega)$, and hence before the time of first difference, it follows that $X$ and $Y$ are also executed at those same times in $\sigma(\tilde{s}, \omega)$. Thus, regardless of the value of $p$, the constraint $Y - X \leq v$ is satisfied, contradicting the choice of $(s, \omega)$.

For the second part, consider the following constraints:

- $C_1$: $(Y - X \leq v,\ \beta\gamma p)$
- $C_2$: $(Y - X \leq v,\ \alpha\beta\gamma)$
- $C_{1.1}$: $(Y - X \leq v,\ \alpha\beta\gamma p)$
- $C_{1.2}$: $(Y - X \leq v,\ (\neg\alpha)\beta\gamma p)$

$C_1$ is the constraint from $X$ to $Y$ shown in Fig. 4. $C_2$ is the constraint derived in the first part of this proof. Now, $C_1$ is equivalent to the pair of constraints, $C_{1.1}$ and $C_{1.2}$, since $\beta\gamma p \equiv (\alpha\beta\gamma p) \vee ((\neg\alpha)\beta\gamma p)$. Thus, the constraint set $\{C_1, C_2\}$ is equivalent to the constraint set $\{C_{1.1}, C_{1.2}, C_2\}$. However, since the label on $C_2$ is subsumed by the label on $C_{1.1}$, the constraint $C_2$ *dominates* the constraint $C_{1.1}$. Thus, the constraint set $\{C_1, C_2\}$ is equivalent to $\{C_{1.2}, C_2\}$. $\square$

This and other label-modification rules are expected to play an important role in the dynamic controllability-checking algorithm that is a major goal of this work.

## Conclusions

This paper presented a temporal network, called a CSTNU, that generalizes CSTPs and STNUs from the literature. The semantics of dynamic controllability for CSTNUs also generalizes the related notions for CSTPs and STNUs. The motivation for this work was to provide a framework for the temporal constraints underlying workflows for business and medical-treatment processes. In future work, we aim to show that any workflow is history-dependent controllable if and only if its underlying CSTNU is dynamically controllable.